\documentclass[conference]{IEEEtran}
\IEEEoverridecommandlockouts
\usepackage{cite}
\usepackage{amsmath,amssymb,amsfonts}
\usepackage{algorithmic}
\usepackage{graphicx}
\usepackage{textcomp}
\usepackage{xcolor}

\usepackage{float} 
\usepackage{url}

\def\BibTeX{{\rm B\kern-.05em{\sc i\kern-.025em b}\kern-.08em
    T\kern-.1667em\lower.7ex\hbox{E}\kern-.125emX}}
\begin{document}

\title{Interactive Image Selection and Training for Brain Tumor Segmentation Network
\thanks{This is a manuscript. The authors thank FAPESP (2023/14427-8, and 2023/09210-0) and CNPq (304711/2023-3) for financial support.
}
}

\author{\IEEEauthorblockN{Matheus A. Cerqueira\IEEEauthorrefmark{1},
Flávia Sprenger\IEEEauthorrefmark{2},
Bernardo C. A. Teixeira\IEEEauthorrefmark{2}\IEEEauthorrefmark{3}, and
Alexandre Falcão\IEEEauthorrefmark{1}}
\IEEEauthorblockA{\IEEEauthorrefmark{1}Institute of Computing, University of Campinas, Campinas, São Paulo, Brazil}
\IEEEauthorblockA{\IEEEauthorrefmark{2}Hospital de Clínicas,  Universidade Federal do Paraná, Curitiba, Paraná, Brazil}
\IEEEauthorblockA{\IEEEauthorrefmark{3}Instituto de Neurologia de Curitiba, Curitiba, Paraná, Brazil}
\thanks{ Corresponding author: M. Cerqueira (email: matheus.cerqueira@students.ic.unicamp.br)}
}

\maketitle

\begin{abstract}

Medical image segmentation is a relevant problem, with deep learning being an exponent. However, the necessity of a high volume of fully annotated images for training massive models can be a problem, especially for applications whose images present a great diversity, such as brain tumors, which can occur in different sizes and shapes. In contrast, a recent methodology, Feature Learning from Image Markers (FLIM), has involved an expert in the learning loop, producing small networks that require few images to train the convolutional layers. In this work, We employ an interactive method for image selection and training based on FLIM, exploring the user's knowledge. The results demonstrated that with our methodology, we could choose a small set of images to train the encoder of a U-shaped network, obtaining performance equal to manual selection and even surpassing the same U-shaped network trained with backpropagation and all training images.

\end{abstract}


\section{Introduction}

Gliomas are the most common type of brain tumor in adults, with the Glioblastoma (GBM) being the most common malignant brain tumor of the Central Nervous System. In 2019 in the United States (US) the survival rate within five years after diagnosis was only 6.9\%, with an incidence rate of 2.55 per 100,000 people~\cite{ostrom2022cbtrus}.

The use of images is important for the initial diagnosis, with volume estimation essential for monitoring, investigating tumor progression, and analyzing the selected treatment~\cite{dupont2016image}. However, manual annotation is time-consuming, tedious, and error-prone -- facts that have motivated research on automatic and semi-automatic methods for brain tumor segmentation.


From Magnetic Resonance Imaging (MRI) sequences, two are the most used to observe the brain sub-regions: Fluid Attenuated Inversion Recovery (T2-FLAIR or simply FLAIR) and the post-gadolinium-based contrast administration T1 (T1Gd). GBMs generally have an irregular shape and size, with active vasogenic edema (ED) on FLAIR and the enhancing tumor (ET) highlighted on T1GD. In addition to ED and ET, a third sub-region can also be observed, the necrotic core (NC), typically as a non-active region in T1Gd, delimited by ET.


Deep Learning (DL) presents the best results among automatic Brain Tumor Segmentation (BTS) techniques. However, traditional DL training requires a high volume of fully-labeled images to train the massive networks and different appearances of tumors. 



Another factor that impacts a dataset's visual appearance is samples from mixed locations acquired from different machines and configurations, such as slice thickness. Active learning is one technique that tries to solve the problem of finding the minimum set of training images\cite{wu2022survey, mosqueira2023human}. However, the process is usually done with an already predefined model without relating to visual characteristics or criteria for such selection, for example, sampling images based on latent representations.


One way to make the process more interesting is to reduce the gap between the user knowledge and the learning loop, such as selecting images. However, to minimize the subject aspect of that interaction, it is essential to have a recommendation based on objective criteria \cite{zhao2021context}.

Therefore, the present work proposes a way of selecting images at the same time that we learn convolutional filters, differing from image selection methods such as active learning. We use the Feature Learning from Image Markers (FLIM) methodology, in which the user draws markers on the images, and the filters are learned directly from these marked regions~\cite{de2020learning,de2020feature}.


We propose an interactive methodology by selecting an image, learning filters with FLIM, and selecting another image that fails according to already learned filters. Our results demonstrate that our data selection obtains results consistent with manual selection and outperforms the results of the model trained with all images of the training set.
    
FLIM differs from traditional scribble learning methods. Traditional methods typically use a pseudo-labeling from the scribbles using a graph method \cite{scribblesup,can2018learning}, for example, or a regularized loss\cite{scribble_reg_loss,scribble_domain}. The fact is that in both cases, the problems related to backpropagation continue to impact those models. On the other hand, the FLIM learning process is direct and does not require a backpropagation algorithm, taking the expert's knowledge into account.

\section{Related Works}


\subsection{Image Selection}

As said before, some works use the user only as the oracle of the annotation, where there is a mechanism for recommending or sampling data, and the user only annotates those samples without properly selecting them. For example, some works measure uncertainty as a Bayesian problem using a probabilistic model \cite{2020_ie3_t_know_delo, 2020_miccai_dd_Al},  and others estimate uncertainty using distances from data representations \cite{2020_miccai_dd_Al, 2018_icmla_med_feat_space}.

On the other hand, some works brought more relevance to the user, closing the gap between selection and annotation. For example, in \cite{2019_ispa_explanation}, the authors pursued ways of recommending data linked to visual explanation, even if the user is still only in the annotation process. In others, the user is the basis of selecting and annotating the data, selecting the data according to specific criteria\cite{zhao2021context}.  However, most of those works are related to training the entire network on each interaction.


\subsection{Feature Learning from Image Markers}

FLIM's previous works show that it is possible to use a reduced number of weakly labeled images to learn a shallow feature extractor (1-3 layers) with a descriptive procedure while maintaining its performance compared to standard deep learning models. It reduces the human effort to mark representative class regions in fewer images. With each marked region as a candidate filter, FLIM learns convolutional filters directly from those marked regions.

However, most works use visual inspection for the image selection method, which can be subjective and time-consuming~\cite{sousa2021cnn,de2022flim_u,cerqueira2023building}. Others used clustering methods and direct 2d projection of images but did so on 2d image datasets for classification and without extracting features from such images\cite{benato2020convolutional,de2020learning}.

\section{Method}

\begin{figure}[H]
    \centering
    \includegraphics[width=0.8\linewidth]{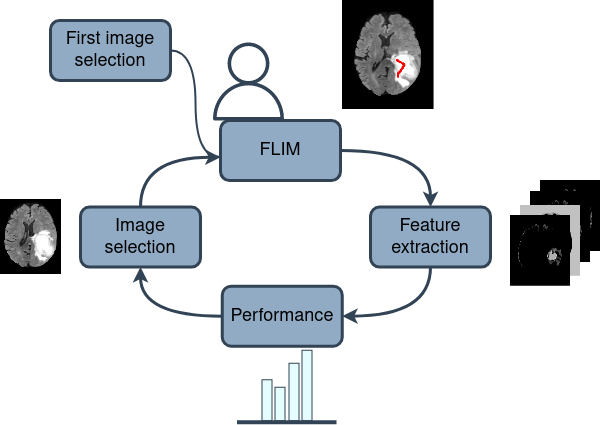}
    \caption{Our interactive methodology. Learn filters from image markers, then select the next image based on the performance of the learned filters.}
    \label{fig:method}
\end{figure}

Our methodology followed the process in Fig. \ref{fig:method}, where the user selects a first image, then marks relevant regions of the image and generates convolutional filters for the network encoder. Such filters are applied to the remaining training images, and then a criterion is applied to obtain the performance of each remaining training image for these existing filters. Finally, in the next step, the user can select an image again, but now selecting the image with the worst performance given the established criteria.

It is worth mentioning that during the learning process, the user annotates the convolutional filters between good filters for WT and ET to compute the criterion based on those regions. Finally, the user can repeat the image selection loop until all images perform well or set a limit of images with markers. In this work, we limited the number of images to 8 for a comparison with \cite{cerqueira2023building}.

Furthermore, it is worth mentioning that we employ the interactive image selection only for the first step, using the already selected images and image markers from the FLIM step to train the other layers of the network encoder.

\begin{figure}[H]
    \centering
    \includegraphics[width=0.83\linewidth]{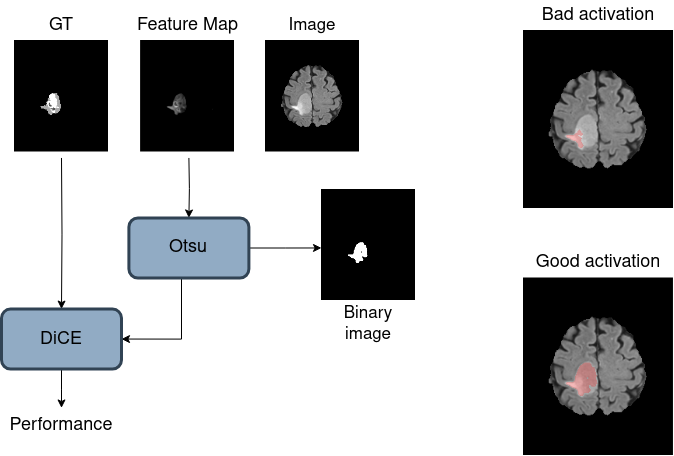}
    \caption{Image selection criteria and examples of good and bad activation superimposed on the original image.}
    \label{fig:criteria}
\end{figure}

Fig. \ref{fig:criteria} presents the criteria used in the selection performance from a query image and an activation map from one learned filter. We compute the binarization of the activation map by using the Otsu threshold. Then, the performance is measured by the Dice score between the ground truth (GT) and the binary image.

Fig. \ref{fig:criteria} also presents examples of two activations (after the binarization), a 'bad' and a 'good' activation. Notice that the first one misses parts of the tumor, and then by selecting this image and learning filters from that, we get the second activation (good), capturing a more significant part of the tumor.

\section{Experiments}

\subsection{Datasets}


We used two datasets, the first is a private dataset containing 80 3D images of GBM (HGG) with two MRI scans (FLAIR and T1Gd) per patient. Each scan was acquired as a volumetric image interpolated to an isotropic resolution of $1mm^3$ and we used the same preprocessing pipeline of \cite{cerqueira2023building}. Also, as a second dataset, we used the BraTS 2020 training dataset, using the FLAIR-T1Gd pair, and as preprocessing, we adopted the histogram normalization mentioned above.

We randomly divided the private dataset into 60\% for training, 10\% for validation, and 30\% for testing. We kept the same amount of training data (50) for the BraTS dataset and separated the remainder between validation and testing (10/90\%). We separated ourselves in this way to have a large set for testing, aiming to check whether the selection of images used can generalize well to the rest of the set.

\subsection{Adopted Architecture}


The sU-Net architecture (\cite{cerqueira2023building}) consists of two encoders, one for T1Gd and the other for FLAIR images, with three convolutional layers each. Skip connections concatenate the output feature blocks, before each strided pooling operation, for both the T1Gd and FLAIR encoders, and in the final layer, a convolution with kernels $1^3$ generates four channels, one for the background and one for each label (ED, ET, NC).

\subsection{Encoder and Decoder training}

We use two learning methods: FLIM to train the network encoder and standard backpropagation to train the decoder. Thus, among the 50 training images, we selected 8 images using the interactive process of figure \ref{fig:method} and the rest of the 50 training images to train the decoder.

We used the exact configuration of data split, learning rate ($2.5e^{-3}$ with linear decay), loss (average of Cross-Entropy and Dice), and a total of 100 epochs. We also used ADAM optimizer and a batch size equal to one.

\subsection{Evaluation Metrics}

We evaluate tumor segmentation into three regions: ET, Tumor Core (TC) and Whole Tumor (WT). The literature usually reports the segmentation effectiveness for these three regions, assuming that $\rm{WT} = \rm{ED} \cup \rm{ET} \cup \rm{NC}$ and $\rm{TC} = \rm{ET} \cup \rm{NC}$. We used the Dice Similarity Coefficient (DSC) to measure efficacy. 

\subsection{Golden Standard Models}

DeepMedic~\footnote{\url{https://github.com/deepmedic/deepmedic} } and nnU-Net~\footnote{\url{https://github.com/MIC-DKFZ/nnUNet} } models were used as golden standard models. These models adopt data augmentation, normalization, and learning rate reduction, providing us with upper-bound metrics. DeepMedic is a dual-branch network that has been shown to use small amount of memory while maintaining performance~\cite{kamnitsas2017efficient}, and nnU-Net is a very relevant network, winning segmentation challenges of the last two years~\cite{isensee2021nnu_brain_tumor, luu2022extending_nnunet}.

\section{Results}


Table \ref{tab:results_ours_vs_base} presents the results of the sU-Net model with different image-selecting methods, either using all training images with standard backpropagation (Backprop.), using FLIM with the user manually selecting the most diverse images for marking ($FLIM_m$), and using the proposed interactive method ($FLIM_i$). It is worth mentioning that the methods that used FLIM froze the encoder, so only the decoder was trained using backpropagation.

The table shows that the interactive method obtained the best mean values and lowest standard deviation, demonstrating the proposed method to select a diverse sub sample of images for training. The interactive method saves the user time from manually selecting those images, and also our methodology based on FLIM outperforms the encoder trained with all training sets using backpropagation.

\begin{figure}[H]
    \centering
    \includegraphics[width=.85\linewidth]{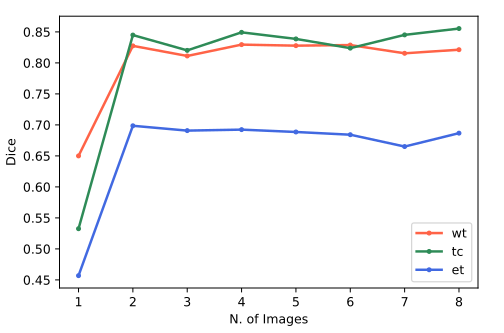}
    \caption{Model performance based on the number of encoder's training images.}
    \label{fig:perform_gbm}
\end{figure}

Furthermore, we verified the model's performance when selecting new images, as shown in Fig. \ref{fig:perform_gbm}, which presents the average Dice between classes for the number of selected images. Note that there is a significant improvement when adding the second image (the first image is recommended). Also, for images 3-8, there is no significant increase in the model's performance, which can be due to the first image being very typical and the second being very difficult, so the gains with the following images were small.

\begin{figure}[H]
  \begin{center}
    \begin{tabular}{ccc} 
      \includegraphics[width=0.2\linewidth]{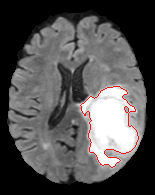} & \includegraphics[width=0.2\linewidth]{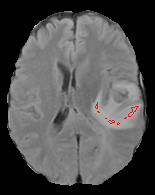} & \includegraphics[width=0.2\linewidth]{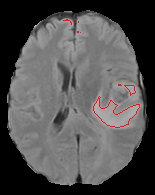} \\
      a & b & c \\
    \end{tabular}
  \end{center}
  \caption{Example of sample and binary image from their best feature:(a) first image, second image with a bad (b)  and good feature (c).}
  \label{fig:good_bad}
\end{figure}

Fig. \ref{fig:good_bad} shows the images from the first and second selections, with the highlighted regions corresponding to the active regions for WT features. In (a), we have the image used on the first selection ($i=1$) and its best feature; in (b), we have the image $i=2$ but with the best feature from the first -- which does not correctly capture the tumor, indicating why this image is recommended. In (c), the same image after training with FLIM is in $i=2$. Note how there is better attention to the tumor in (c), which corresponds with the improvement in the final image prediction, going from a Dice score of $0.01$ to $0.64$ in the entire model, which corresponds to the improvement observed in the selection criterion, which was from $0.15$ to $0.56$.

Thus, We can correlate the final image metrics with its performance in the developed image selection criteria and with the features learned in the first layer, which brings security to the developed model. Unfortunately, our criteria use the GT of the image, which prevents us from obtaining a reliable measure when making an inference from the test image that does not have GT. Otherwise, this would be an excellent tool for using a system in clinical environments, providing not only the segmentation mask but also to which features it is related. 


Next, we compare our trained model with the gold standard state-of-the-art models (Table \ref{tab:results_ours_vs_sota}, note that our goal is not to beat such models, since we use a much leaner network that trains fast, but rather to obtain an estimate of how close (far) our results are to such massive networks. Here, we see that nnU-Net performed better, as expected. However, our results are close to such models, even using around 3\% number of nnU-Net parameters.

\begin{table}[h]
\caption{Evaluation metrics of multiple image selection method: No image selection (Backprop.),  FLIM with manual selection ($FLIM_m$), and FLIM with interactive selection ($FLIM_i$)}
\label{tab:results_ours_vs_base}
\begin{center}       
\begin{tabular}{|l|l|l|l|}
\hline
\rule[-1ex]{0pt}{3.5ex}  Models & \multicolumn{3}{|c|}{DSC $\uparrow$} \\
\hline
\rule[-1ex]{0pt}{3.5ex}    & ET & TC & WT \\
\hline
\rule[-1ex]{0pt}{3.5ex}  Backprop. & 0.665 $\pm$ 0.166 & 0.734 $\pm$ 0.157 & 0.721 $\pm$ 0.104  \\
\hline
\rule[-1ex]{0pt}{3.5ex} $FLIM_m$ & 0.691 $\pm$  0.073 & 0.733 $\pm$  0.072 & 0.702 $\pm$ 0.109 \\
\hline
\rule[-1ex]{0pt}{3.5ex}  $FLIM_i$ & \textbf{0.713 $\pm$ 0.068}  & \textbf{0.810 $\pm$ 0.066} & \textbf{0.797 $\pm$ 0.065}  \\ 
\hline
\end{tabular}
\end{center}
\end{table}

\begin{table}[h]
\caption{Evaluation metrics of our method against SOTA models for the GBM dataset.}
\label{tab:results_ours_vs_sota}
\begin{center}       
\begin{tabular}{|l|l|l|l|}
\hline
\rule[-1ex]{0pt}{3.5ex}  Models & \multicolumn{3}{|c|}{DSC $\uparrow$}  \\
\hline
\rule[-1ex]{0pt}{3.5ex}    & ET & TC & WT  \\
\hline
\rule[-1ex]{0pt}{3.5ex}  DeepMedic & 0.777 $\pm$ 0.056  & 0.851 $\pm$ 0.066 & 0.792 $\pm$ 0.094  \\ 
\hline
\rule[-1ex]{0pt}{3.5ex}  nnU-Net & \textbf{0.798 $\pm$ 0.045} & \textbf{0.885 $\pm$ 0.058} & \textbf{0.851 $\pm$ 0.068}  \\
\hline 
\rule[-1ex]{0pt}{3.5ex}  Ours & 0.713 $\pm$ 0.068  & 0.810 $\pm$ 0.066 & 0.797 $\pm$ 0.065  \\
\hline
\end{tabular}
\end{center}
\end{table}

\begin{table}[h]
\caption{Evaluation metrics of our method against SOTA models for the BraTS dataset.}
\label{tab:results_ours_vs_sota_brats}
\begin{center}       
\begin{tabular}{|l|l|l|l|}
\hline
\rule[-1ex]{0pt}{3.5ex}  Models & \multicolumn{3}{|c|}{DSC $\uparrow$}  \\
\hline
\rule[-1ex]{0pt}{3.5ex}    & ET & TC & WT  \\
\hline
\rule[-1ex]{0pt}{3.5ex}  DeepMedic & 0.777 $\pm$ 0.175  & 0.810 $\pm$ 0.196 & 0.808 $\pm$ 0.138 \\ 
\hline
\rule[-1ex]{0pt}{3.5ex}  nnU-Net & \textbf{0.842 $\pm$ 0.153} & \textbf{0.884 $\pm$ 0.163} & \textbf{0,906 $\pm$ 0.089}   \\
\hline 
\rule[-1ex]{0pt}{3.5ex} Ours &  0,717 $\pm$ 0,223   & 0,733  $\pm$ 0,237 &  0,789 $\pm$ 0,184 \\
\hline
\rule[-1ex]{0pt}{3.5ex} Backprop. &  0,717 $\pm$ 0,214   & 0,734  $\pm$ 0,239 &  0,772 $\pm$ 0,184 \\
\hline
\end{tabular}
\end{center}
\end{table}

\section{Conclusion}

Finding the smallest set of images that efficiently trains a network is a challenge. In the present work, we use a methodology that selects the training images while obtaining the convolutional filters from the encoder. The user draws markers on the selected images, learning convolutional filters from such markers. Then, the following training images can be selected according to the performance of the already learned filters. As a result, we selected a small set of images that trained the encoder of a U-shaped network, obtaining performance similar to manual selection and surpassing the performance of the network trained with all available images. We wish to use the methodology for images of other natures in future work.





\bibliographystyle{IEEEtran}

\bibliography{example}

\end{document}